\documentclass[10.5pt,compsoc]{tst}
\pdfoutput=1
\usepackage{makecell}
\usepackage{graphicx}
\usepackage{footmisc}
\usepackage{subfigure}
\usepackage{url}
\usepackage{multirow}
\usepackage[noadjust]{cite}
\usepackage{amsmath,amsthm}
\usepackage{amssymb,amsfonts}
\usepackage{booktabs}
\usepackage{color}
\usepackage{ccaption}
\usepackage{booktabs}
\usepackage{float}
\usepackage{fancyhdr}
\usepackage{caption}
\usepackage{xcolor,stfloats}
\usepackage{comment}
\graphicspath{{figures/}}
\usepackage{cuted}  
\usepackage{captionhack}
\usepackage{epstopdf}
\usepackage{tabularx}

\headevenname{\normalsize{\textbf{\emph{Tsinghua Science and Technology, xxxxx}} 20xx, 2x(x): xxx--xxx}}%
\headoddname{{\sf }\quad {\textbf{\emph{Joint Extraction of Uyghur Medicine 
Knowledge with Edge Computing}}}}%

\newtheoremstyle{mystyle}{0pt}{0pt}{\normalfont}{1em}{\bf}{}{1em}{}
\theoremstyle{mystyle}
\renewcommand\figurename{Fig.~}

\newcommand{\upcite}[1]{\textsuperscript{\cite{#1}}}

\newcommand{\nop}[1]{}

\makeatletter
\renewcommand{\@biblabel}[1]{[#1]\hfill}
\makeatother

\begin{document}

\hyphenpenalty=50000

\makeatletter
\newcommand\mysmall{\@setfontsize\mysmall{7}{9.5}}

\newenvironment{tablehere}
  {\def\@captype{table}}
  {}
\newenvironment{figurehere}
  {\def\@captype{figure}}
  {}

\thispagestyle{plain}%
\thispagestyle{empty}%

\let\temp\footnote
\renewcommand \footnote[1]{\temp{\normalsize #1}}
{}
\vspace*{-40pt}
\noindent{\normalsize\textbf{\scalebox{0.885}[1.0]{\makebox[5.9cm][s]
{TSINGHUA\, SCIENCE\, AND\, TECHNOLOGY}}}}

\vskip .2mm
{\normalsize
\textbf{
\hspace{-5mm}
\scalebox{1}[1.0]{\makebox[5.6cm][s]{%
I\hfill S\hfill S\hfill N\hfill{\color{white}%
l\hfill l\hfill}1\hfill0\hfill0\hfill7\hfill-\hfill0\hfill2\hfill1\hfill4
\hfill \color{white}{\quad 0\hfill ?\hfill /\hfill ?\hfill ?\quad p\hfill p\hfill  ?\hfill ?\hfill ?\hfill --\hfill ?\hfill ?\hfill ?}\hfill}}}}


\vskip .2mm\noindent
{\normalsize\textbf{\scalebox{1}[1.0]{\makebox[5.6cm][s]{}}}}%

\begin{strip}
{\center
{\LARGE\textbf{
Joint Extraction of Uyghur Medicine 
Knowledge with Edge Computing}}


\vskip 9mm}

{\center {\sf \large
Fan Lu, Quan Qi$^*$, Huaibin Qin
}
\vskip 5mm}

\centering{
\begin{tabular}{p{160mm}}

{\normalsize
\linespread{1.6667} %
\noindent
\bf{Abstract:} {\sf
Edge computing, a novel paradigm for performing computations at the network edge, holds significant relevance in the healthcare domain for extracting medical knowledge from traditional Uyghur medical texts. Medical knowledge extraction methods based on edge computing deploy deep learning models on edge devices to achieve localized entity and relation extraction. This approach avoids transferring substantial sensitive data to cloud data centers, effectively safeguarding the privacy of healthcare services. However, existing relation extraction methods mainly employ a sequential pipeline approach, which classifies relations between determined entities after entity recognition. This mode faces challenges such as error propagation between tasks, insufficient consideration of dependencies between the two subtasks, and the neglect of interrelations between different relations within a sentence. To address these challenges, a joint extraction model with parameter sharing in edge computing is proposed, named CoEx-Bert. This model leverages shared parameterization between two models to jointly extract entities and relations. Specifically, CoEx-Bert employs two models, each separately sharing hidden layer parameters, and combines these two loss functions for joint backpropagation to optimize the model parameters. Additionally, it effectively resolves the issue of entity overlapping when extracting knowledge from unstructured Uyghur medical texts by considering contextual relations. Finally, this model is deployed on edge devices for real-time extraction and inference of Uyghur medical knowledge. Experimental results demonstrate that CoEx-Bert outperforms existing state-of-the-art methods, achieving accuracy, recall, and F1 scores of 90.65\%, 92.45\%, and 91.54\%, respectively, in the Uyghur traditional medical literature dataset. These improvements represent a 6.45\% increase in accuracy, a 9.45\% increase in recall, and a 7.95\% increase in F1 score compared to the baseline.}
}

\end{tabular}
}
\vskip 6mm

\vskip -3mm
\small\end{strip}


\begin{figure}[b]
\vskip -6mm
\begin{tabular}{p{44mm}}
\toprule\\
\end{tabular}
\vskip -4.5mm
\noindent
\setlength{\tabcolsep}{1pt}
\begin{tabular}{p{1.5mm}p{79.5mm}}

$\bullet$& Fan Lu, Quan Qi and Huaibin Qin are with the School of Information Science and Technology, Shihezi University, Shihezi 832000, China. E-mail: fanbyxc1014@gmail.com; Q.Qi@ieee.org; hbqin1023@163.com. \\

$\sf{*}$&
To whom correspondence should be addressed. \\
          &          Manuscript received: 2023-09-20;
          revised1: 2023-11-8;
          revised2: 2023-11-30;
          accepted: 2023-12-23

\end{tabular}
\end{figure}\large

\vspace{3.5mm}
\section{Introduction}
\label{s:introduction}
\noindent

The expanding field of medicine, driven by advanced research and innovative clinical practices, has led to an exponential growth of medical literature. Extracting key information and insights from these resources has become crucial\upcite{1}. Uyghur medicine, an integral part of Chinese traditional culture, possesses unique medical customs that have gained the trust and admiration of the Uyghur population and other ethnic minorities. It is characterized by its autonomous theoretical framework, vast array of medicinal resources, and distinctive diagnostic and treatment paradigms\upcite{2}. Applying text extraction technologies to Uyghur medical literature is extremely relevant, as it not only preserves and uncovers the rich Uyghur medical culture but also integrates Uyghur medicine into mainstream clinical practices\upcite{3}. By extracting and analyzing Uyghur language medical literature, including prescriptions, medication details, and treatment strategies, unique drug utilization patterns and therapeutic outcomes can be revealed. This provides valuable insights and directions to healthcare professionals, thus diversifying the landscape of clinical practice\upcite{4}.

Traditional extractions of Uyghur medical entities and relations commonly employ rule-based or template matching approaches. These manual processes typically involve creating a Uyghur language medical dictionary that includes common entity names such as herbs, diseases, symptoms, and prescriptions, along with their respective part-of-speech tags and lexical meanings\upcite{5}. Then, regular expressions, grammatical rules, or template matching principles are established based on Uyghur syntactic and semantic properties to extract relevant entities and relations from raw text materials. These extracted entities and relations are further classified and annotated. However, these conventional techniques for extracting Uyghur medicinal entities and relations have inherent limitations:

\setlength{\hangindent}{18pt}
\noindent

  $\bullet$ \textbf{Reliance on manual rule design and annotation}: Traditional techniques for entity and relation extraction require manual creation of vocabulary, establishment of regular expressions, grammatical rules, and the development of template matching standards, as well as manual annotation. This method consumes significant human, monetary, and temporal resources\upcite{6}.

\setlength{\hangindent}{18pt}
\noindent

$\bullet$ \textbf{Annotation bias and inconsistency}: Differences in annotation standards and individual annotators cognitive perspectives often lead to variation and inconsistency in entity and relation annotations. These variations can compromise the accuracy and uniformity of extraction outcomes.

\setlength{\hangindent}{18pt}
\noindent

  $\bullet$ \textbf{Limited coverage}: Conventional techniques for extracting entities and relations are dependent on pre-established rules, dictionaries, and templates, making them inherently constrained. They have limited adaptability when encountering novel entities and relations that were not considered during the rule establishment phase.

In recent years, edge computing has emerged as a significant research focus and has garnered considerable attention\upcite{7}. Edge computing facilitates localized computation processes near the data source, aiming to alleviate network congestion and enhance real-time processing capability\upcite{8}. Edge computing represents a distributed computing paradigm that moves data processing and analytics away from traditional centralized models towards edge devices in close proximity to the data source\upcite{9}. This shift is primarily driven by the demands of big data, the Internet of Things (IoT), and real-time applications, as conventional cloud computing models may face challenges in latency and bandwidth when meeting these needs\upcite{10}.

While edge computing offers remarkable advantages, it also presents certain challenges, particularly in terms of security and privacy vulnerabilities. Distributing data processing across edge devices can potentially expose data to risks of leakage and malicious intrusions\upcite{11}. Consequently, researchers and tech companies are striving to develop secure edge computing solutions to ensure data confidentiality and integrity. In essence, edge computing epitomizes an emerging computing paradigm that addresses novel demands in data processing\upcite{12}. This transition has profound impacts on our technological interactions across various domains. In the field of medicinal knowledge extraction, edge computing brings numerous enhancements and benefits, particularly in the extraction of Uyghur medicinal knowledge\upcite{13}. Here are some key roles played by edge computing in this area:

\setlength{\hangindent}{18pt}
\noindent

  $\bullet$ \textbf{Real-time Extraction and Inference}: Edge computing strategically deploys computational resources closer to the data source, enabling real-time extraction and inference of medical knowledge\upcite{14}.

\setlength{\hangindent}{18pt}
\noindent

$\bullet$ \textbf{Data Privacy and Security}: Traditional data transmission carries inherent risks, especially when it involves medical data. Edge computing localizes the deployment of models on devices, eliminating the need for sensitive data transmission. This reduces the risk of data breaches and protects the privacy of medical knowledge.

\setlength{\hangindent}{18pt}
\noindent

$\bullet$ \textbf{Low Latency}: The extraction of medical knowledge often requires processing large volumes of unstructured text data. Edge computing, with its localization characteristic, eliminates the latency associated with bidirectional data transfer to remote, centralized servers. This improves extraction efficiency and accuracy.

\setlength{\hangindent}{18pt}
\noindent

  $\bullet$ \textbf{Offline Usability}: Edge computing enables the operation of extraction models in scenarios with unstable network connectivity, such as remote areas or environments with unreliable networks. This allows the extraction and application of Uyghur medicinal knowledge even in offline environments.

In conclusion, edge computing offers several benefits, including real-time processing, enhanced security, and reduced latency, for extracting Uyghur medicinal knowledge. These benefits form a strong foundation for research and application in medicinal knowledge extraction. By integrating deep learning models with edge computing technologies, it is possible to achieve a better understanding and gain more insights from Uyghur traditional medicine. This ultimately promotes innovation in medical research and clinical practice. This paper makes significant research contributions in the following four aspects:

Firstly, it introduces CoEx-Bert, an innovative joint extraction model. Leveraging the power of the pre-trained BERT architecture, the model extracts important features from medical text and utilizes multi-task learning to simultaneously extract various entity relations. CoEx-Bert combines a shared BERT encoder with dedicated output layers specific to certain types, achieving end-to-end joint extraction of entities and relations.

Secondly, the paper conducts a comprehensive series of experiments to validate the performance of the proposed model. Using the Uyghur medical dataset as an empirical foundation and comparing against strong baseline methods, the experiments demonstrate the effectiveness and superiority of the CoEx-Bert model in medical entity-relation extraction.

Thirdly, the model is deployed using an edge architecture, which reduces bandwidth consumption and latency, enabling real-time extraction.

Fourthly, the study evaluates the model using authentic traditional Chinese medical text data, showcasing the reliability and benefits of the algorithm in practical applications, particularly in the healthcare context. The empirical validation highlights the robustness and usefulness of the approach in achieving real-time extraction.

The rest of the paper is structured as follows: Section 2 provides a review of prior research, while Section 3 offers essential background knowledge. Section 4 establishes the extraction model and describes its specific architecture. Section 5 presents the experiments and validation of results, and finally, Section 6 provides a conclusion.

\section{Related Work}
\label{s:Review of Prior Research}
\noindent

In recent years, significant progress has been made in the fields of deep learning and Natural Language Processing (NLP), leading to the widespread application of entity and relation extraction methods based on machine learning. These machine-learning-based methodologies have demonstrated robust capabilities in accurately identifying entities and their respective relations, while also exhibiting adaptability to various natural language expressions.

For example, Zhou et al.\upcite{15} emphasized the improvement of relation extraction accuracy through the combination of Convolutional Neural Networks (CNNs) with Bi-directional Long Short-Term Memory (LSTM) units. They also used a uni-directional LSTM to decode entity pairs corresponding to the predicted relations. This integration of neural components strengthens the robustness of relation extraction. Similarly, Yuan et al.\upcite{16} proposed the relation-specific attention network, which includes a gating mechanism responsible for allocating attention across different aspects of a sentence based on the relation type. This mechanism effectively suppresses the interference of irrelevant relations during the entity recognition process. Joint extraction methods, which utilize cross-task parameter sharing, provide a unified platform for the simultaneous extraction of entities and relations. Sui et al.\upcite{17} introduced a pioneering joint extraction paradigm that uses sequence tagging, supported by an individual decoder, to identify spans corresponding to entities and relations within a sentence. Building on this scheme, Dai et al.\upcite{18} enhanced the process by designing a novel labeling method where a sentence of length $n$ is labeled $n$ times based on the position of each word $w$, allowing for multiple iterative annotations of every word and promoting a more nuanced labeling process.

Miwa et al.\upcite{19} initially proposed a tabular representation framework for structuring entities and relations, transforming the extraction task into a table populating job. However, this method still requires complex feature engineering. Subsequently, Liu et al.\upcite{20} introduced an end-to-end neural network that employs shared parameters to coordinate joint extraction and integrate independent output channels. However, the propagation of redundant information remains a potential issue. Liu et al.\upcite{21} conducted an in-depth study on the method of Chinese word segmentation, proposing an improved algorithm for the conditional field model CRF++ and optimizing extraction results.

In response to the challenges in entity relation extraction from Uighur medical texts, such as error propagation between tasks, insufficient consideration of dependency relationships between two subtasks, and neglect of interrelations among different relationships within a sentence, as well as concerns regarding model deployment in cloud computing centers involving extensive transmission of sensitive data, posing challenges in ensuring privacy and security, this paper proposes a joint extraction method based on the principle of parameter sharing at the edge. This approach utilizes two distinct models to independently extract the principal entity (S) and the combination of the relation + object entity (PO). During the extraction processes of these two models, parameter sharing is employed, summing up the losses of both models to minimize the overall loss of the joint model and conducting backpropagation collaboratively. This results in an end-to-end entity relation extraction process. Compared to conventional single-model entity relations extraction methods, the parameter-sharing-based joint extraction can improve extraction accuracy and efficiency to a certain extent\upcite{22}. Furthermore, this proposed method can better capture contextual information among entities and relations, yielding more robust and accurate extraction results. Additionally, it has the potential to address problems of non-continuity and type confusion, while to some extent preserving the privacy of the data.

\section{Preliminary Knowledge}
\noindent

\subsection{BERT Model}
\noindent

BERT (Bidirectional Encoder Representations from Transformer) is an exceptional language model widely used in various NLP tasks, such as text categorization, named entity recognition, and relation extraction\upcite{23}. The architecture of BERT is shown in Fig.1.

\begin{figure}[h]
  \centering
  \includegraphics[width=7.0cm]{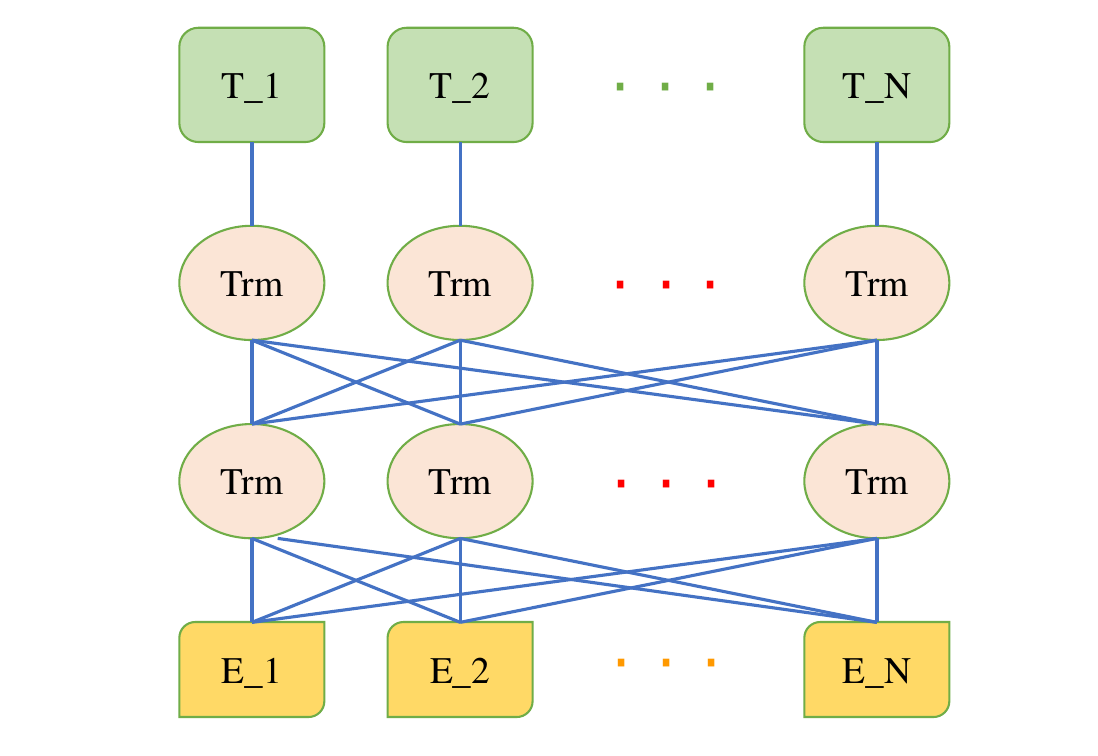}
  \caption{BERT's Multi-Layer Encoder Structure.}
  \label{fig:1}
\end{figure}

The most common approach for relation extraction tasks is the extraction pipeline or pipeline model\upcite{24}. This method processes natural language text through multiple stages to extract structured knowledge, including entities, relations, and events. At each stage, the input text is processed into a specific structure or label before being passed onto the next stage for further processing, eventually resulting in structured knowledge. This method effectively utilizes various NLP techniques to improve the accuracy and efficiency of knowledge extraction.

However, the knowledge extraction pipeline has several limitations. Firstly, the processing at each stage is relatively independent, which may lead to error accumulation and reduce the final extraction accuracy. Secondly, some complex relations and events may require collaborative processing across multiple stages to obtain comprehensive structured knowledge, which pipeline models struggle to support. Finally, as the processing at each stage is fixed, pipeline models find it difficult to adapt to different tasks and languages, requiring specific optimizations and improvements for individual scenarios.

\subsection{Joint Extraction}
\noindent

Joint extraction refers to the concurrent extraction of entities, relations, events, and other structures, as opposed to the staged approach adopted by the knowledge extraction pipeline\upcite{25}. Compared to the knowledge extraction pipeline, joint extraction provides the following advantages:

\setlength{\hangindent}{18pt}
\noindent

  $\bullet$ \textbf{More accurate extraction results}: Joint extraction considers information on entities, relations, and other structures simultaneously. Through joint optimization, it reduces error accumulation and leads to more accurate extraction outcomes\upcite{26}.

\setlength{\hangindent}{18pt}
\noindent

  $\bullet$ \textbf{Better scalability}: The knowledge extraction pipeline operates on a stagewise processing paradigm, which requires modifying other stages when adjusting or adding a processing method at any stage. In contrast, the joint extraction model is more flexible when adapting to new tasks and languages, only requiring the addition of appropriate annotations during model training\upcite{27}.

\setlength{\hangindent}{18pt}
\noindent

  $\bullet$ \textbf{Better generalization ability}: Joint extraction can learn the relations between multiple structures simultaneously and better utilize context information, resulting in improved generalization ability when processing new texts\upcite{28}.

In conclusion, when compared to the knowledge extraction pipeline, joint extraction demonstrates superior precision, efficiency, scalability, and generalizability, as illustrated in Fig.2. It represents one of the most advanced research directions in the current landscape of knowledge extraction.

\begin{figure}[h]
  \centering
  \includegraphics[width=\linewidth]{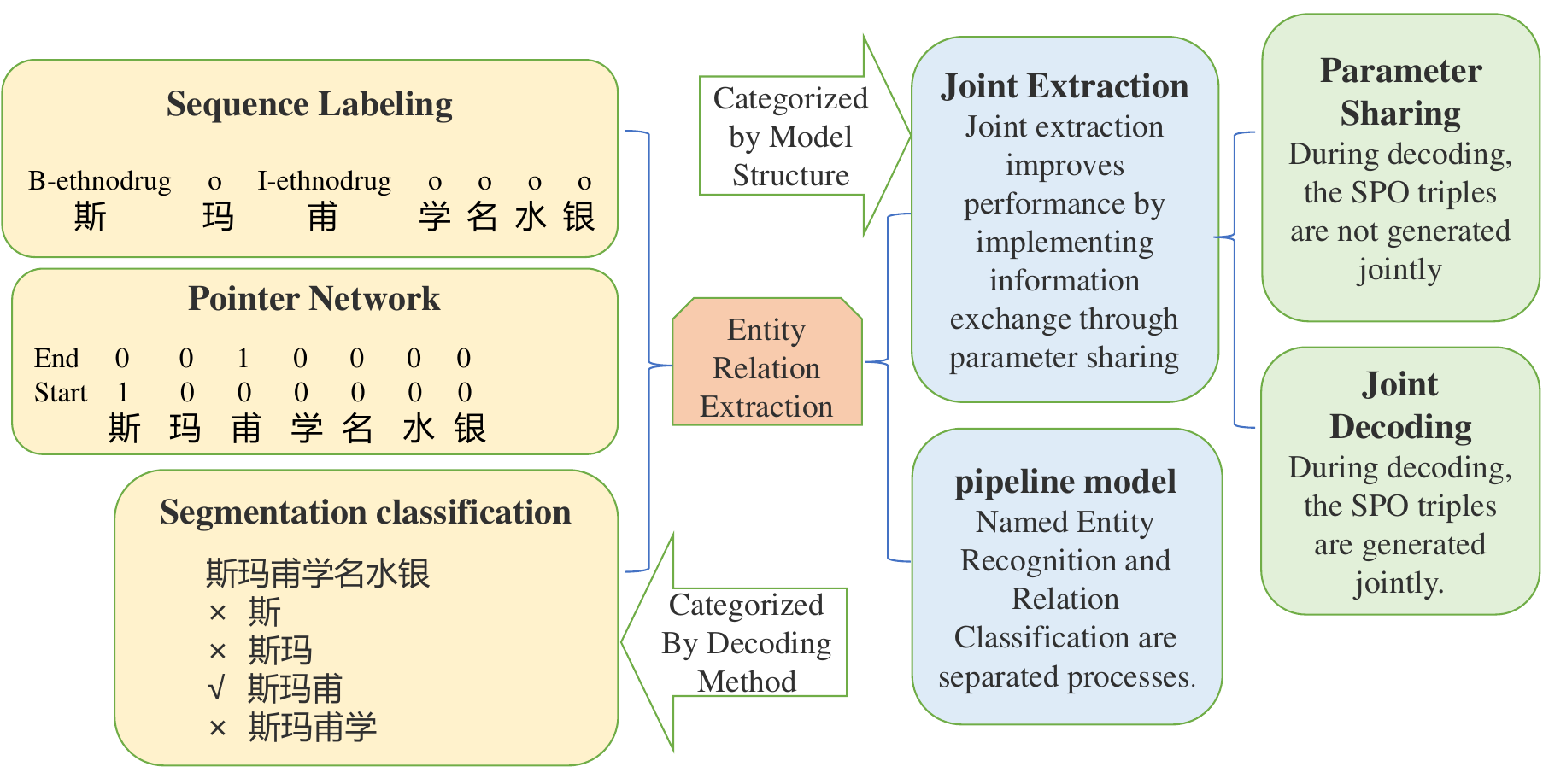}
  \caption{Mainstream Methods of Entity Relation Extraction.}
  \label{fig:2}
\end{figure}

As illustrated in Fig.2, this paper emphasizes the extraction of Chinese text, and this process is demonstrated through an example of Chinese word segmentation. The figure includes examples of Chinese field annotations, which intuitively showcase three decoding classification modes: sequence annotation, pointer annotation, and segmentation classification annotation as three decoding classification modes.

\section{Proposed Methodology}
\noindent

NLP technology is increasingly being utilized for extracting information in the medical field. However, the restricted language and cultural development of Uyghur ethnic medicine pose challenges, resulting in a limited amount of Uyghur-language medical literature. Most medical literature on Uyghur medicine is published in Chinese. Additionally, as the Chinese corpus continues to expand, extracting information and relations from Chinese literature has become common practice. Therefore, it is more convenient and effective to extract information and relations about traditional Uyghur medicines from Chinese literature.

Considering this background, we propose a novel model architecture called CoEx-Bert, based on the BERT structure. First, the input data is embedded to vectorize it. Then, it goes through a Multi-Head Attention(MHA) mechanism layer, which assigns weights according to contextual information. Next, it proceeds to the following layer for Additive Residual Connection (ARC) and Layer Normalization (LN) tasks, addressing issues of gradient vanishing and explosion while standardizing output variables. The standardized parameters are input into a feed-forward neural network for extracting head-tail entity features\upcite{29}. 
Once the extraction is complete, the hidden layer parameters are fed into another model responsible for extracting relations between entities. Both sets of hidden layer parameters are added together positionally and passed into a fully connected layer for classification. The losses from both models are summed up and optimized through backpropagation.

\begin{figure*}[h]
  \centering
  \includegraphics[width=\textwidth]{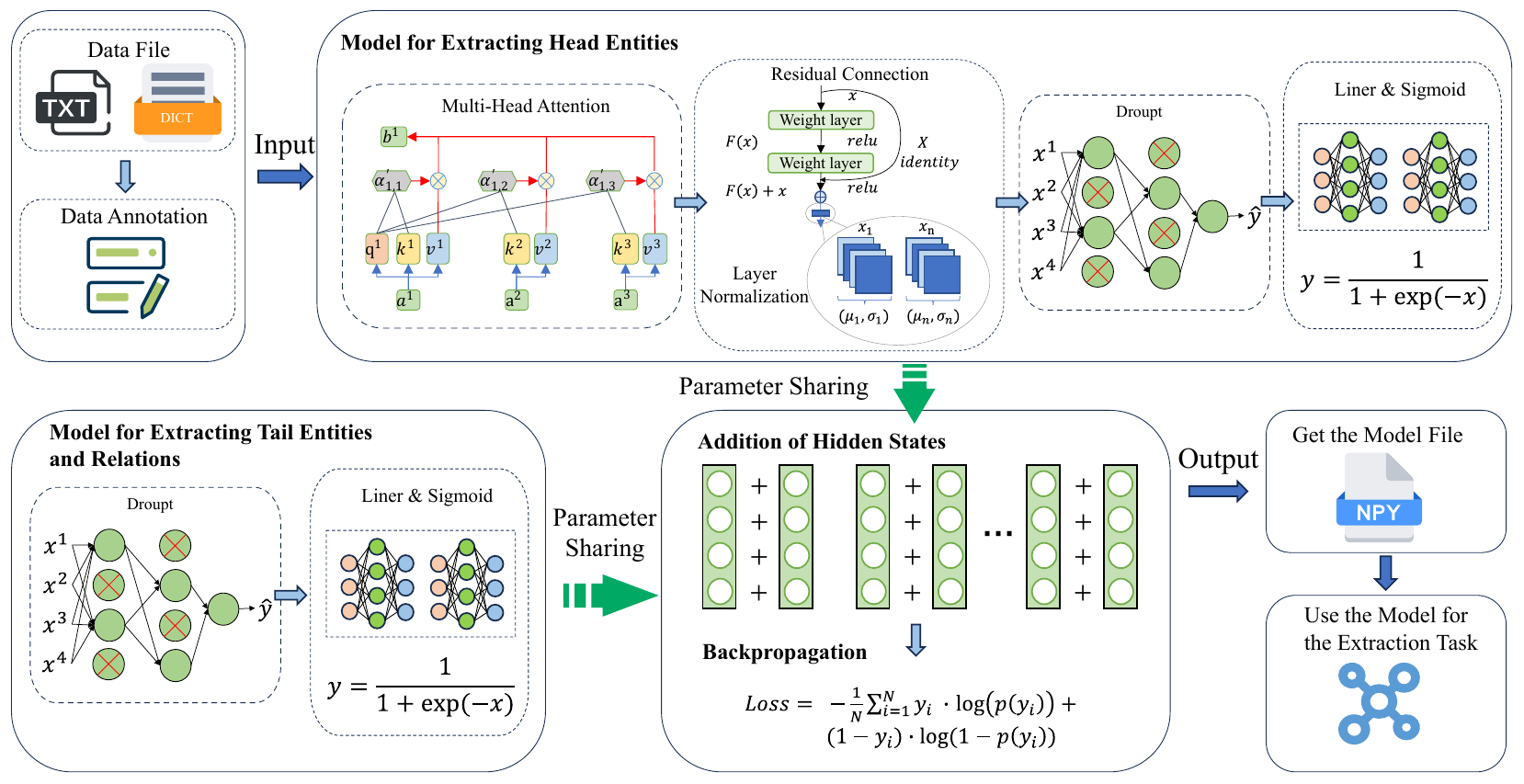}
  \caption{Overall structure diagram of CoEx-BERT.}
  \label{Fig:3}
\end{figure*}

After the extraction of entity features is completed, the hidden layer parameters are passed to another model responsible for extracting relations between entities. Simultaneously, the hidden layer parameters from both models are merged and then passed to a fully connected layer for classification. The collective loss from both models is aggregated and used for optimization based on backpropagation. The overall model architecture is an end-to-end pattern specifically designed for information and relation extraction in the field of Uyghur traditional medicine. Ultimately, it forms a comprehensive framework for extracting information and relations within the domain of Uyghur traditional medicine. The overall model structure can be seen in Fig.3. In the attention mechanism mentioned, each parameter's meaning is described in Table 1.

\begin{table}[h]
\parbox[t][0.5cm]{8cm}{\caption{\textbf{The meaning represented by each parameter.}}}

\label{tab:2}
  \begin{tabular}{c p{6.2cm}}
 \hline
    Notation &   \multicolumn{1}{c} {Definition}  \\
   \hline
  
    $a$  & Result vector after attention weighting  \\

    $q$  & Query vector\\

    $k$  & Key vector  \\

    $v$  & Value vector \\

    $b$  & Bias vector  \\

    $x$  & input vector \\

    $\mu$  & mean value \\

    $\delta$   & standard deviation \\

\bottomrule
\end{tabular}

\end{table}




In the context of LN graphs, subsequent terms clarify the roles of key components. These parameters collectively regulate the normalization of input vectors. Each element within vector $x$ is standardized by subtracting its mean $\mu$ and dividing by its standard deviation $\delta$. This process results in a normalized vector that is suitable for modeling input data while mitigating the impact of possible variations.

\subsection{Overall Structure of the Model}
\noindent

In this paper, we propose the novel joint extraction model CoEx-BERT for extracting diverse medical entities, including diseases, drugs, and symptoms, from Chinese medical texts. The core of this method is a parameter-sharing joint extraction model combined with the pre-trained BERT model, which helps extract important features from medical text data. Additionally, we establish a multi-task learning model that includes two different sub-models working together through parameter sharing. This configuration effectively facilitates the completion of parallel joint extraction tasks.

CoEx-BERT demonstrates the capability to extract various medical entities, such as diseases, drugs, and symptoms, from Chinese medical texts. The approach revolves around a parameter-sharing joint extraction model and the utilization of pre-trained BERT models to extract significant features from medical text data. Furthermore, we establish a multi-task learning model consisting of two distinct sub-models that work cohesively through parameter sharing, resulting in effective parallel joint extraction tasks.

The CoEx-BERT model consists of a sequence of Transformer encoders, each encoder incorporating a MHA mechanism layer and a FFNN. The input to each encoder is derived from the output of its preceding encoder, encapsulated within the framework of ARC and LN. ARC is smartly integrated to merge the input and output, enhancing the model's ability to learn residual information from the input data. Simultaneously, normalization procedures standardize the mean and variance in the network's input vector\upcite{30}. This normalization process harmonizes the dimensions of the model's parameters, leading to smoother training and optimization phases, facilitating effective convergence. The mechanics of how CoEx-Bert skillfully extracts textual information are elucidated in Fig.4, which provides examples in Chinese, offering a more intuitive demonstration of the training process for Chinese text. For Chinese text, CoEx-Bert performs character embedding before initiating the training process.

\begin{figure}[h]
  \centering
  \includegraphics[width=\linewidth]{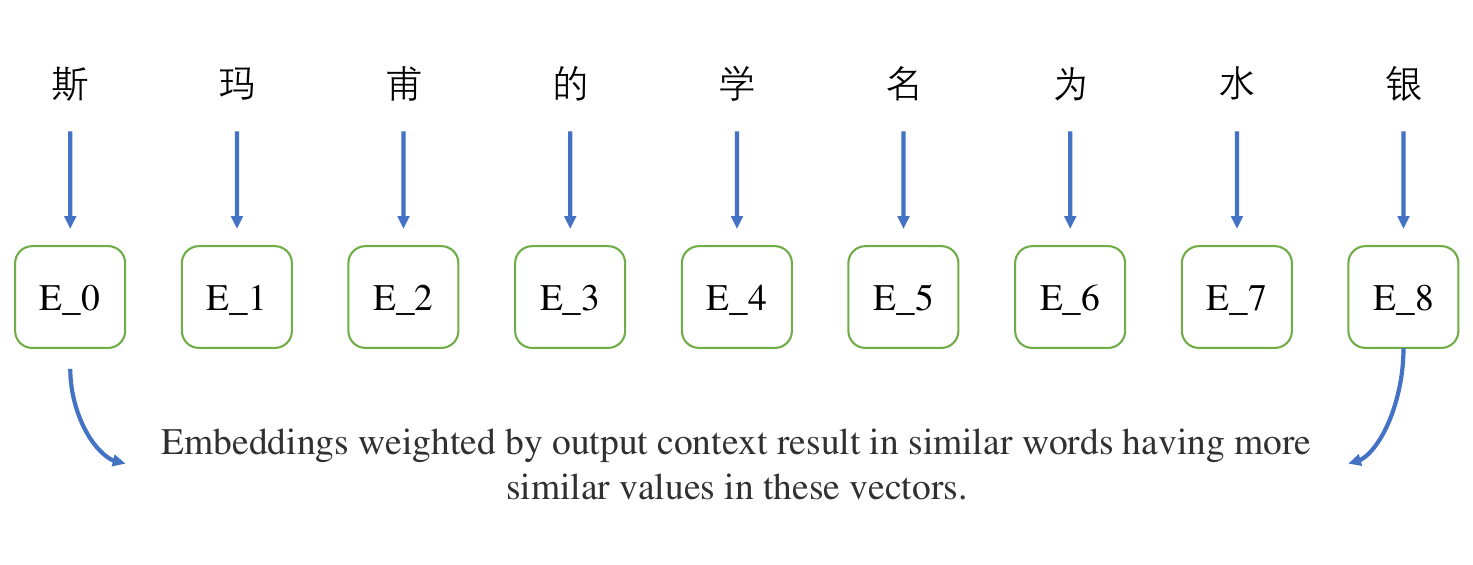}
  \caption{Extracting Textual Information with CoEx-Bert.}
  \label{Fig:4}
\end{figure}

\subsection{Structural Modeling of Each Layer}
\noindent

The structure begins with the MHA layer, where the formula can be broken down into three sections: linear transformations, computation of attention weights, and weighted summation. The input vectors $x$ are first transformed into their corresponding Query ($q$), Key ($k$), and Value ($v$) vectors via multiple transformation matrices $W_Q,W_K,W_V$, as demonstrated in Eq.(1). Following this, the similarity between $q$ and $k$ is calculated, typically using Dot Product or Scaled Dot Product to obtain the attention weights $\alpha$ as shown in Eq.(2). Here, $d_k$ denotes the dimension of vector $k$, division by $d_k$ after dot product application serves a scaling purpose, assisting in avoiding issues of excessively large or small inputs for softmax. Finally, $\alpha$ is applied as a weight metric for $v$ during the weighted summation, resulting in the final output vector $z$, as depicted in Eq.(3).

Within the MHA mechanism, multiple sets of linear transformations, attention weight computations, and weighted summations are typically executed\upcite{31}. Each set transforms the input vector into corresponding Query, Key, and Value vectors via distinct linear operations. These sets then calculate multiple attentions in parallel, and the results are amalgamated in a fixed manner. This approach enables the model to comprehend the correlation between the input sequence and different feature spaces, allowing it to capture long-range dependencies and semantic features inherent in the input sequence, thereby enhancing performance. 

\begin{equation}
q = xW_Q, \ k = xW_K, \ v = xW_V
\end{equation}

\begin{equation}
\alpha = softmax(\frac{qk^T}{\sqrt{d_k}})
\label{eq:2}
\end{equation}

\begin{equation}
z = \alpha v
\label{eq:3}
\end{equation}

Subsequent to this is the FFNN layer. The FFNN, being the most rudimentary form of an artificial neural network, has gained significant traction in its deployment across a wide array of NLP tasks. 

This structure, characterized by interconnected neurons, operates on the principle of forward propagation. Within the FFNN framework, each neuron accepts weighted inputs from its predecessor neurons, processes these inputs via an activation function, and subsequently forwards the result to the neurons in the subsequent layer. This unidirectional signal propagation approach circumvents the occurrence of loops within the neural network. In the realm of NLP tasks, FFNN finds extensive application in classification, labeling, and extraction processes. Its underpinning mathematical model is depicted in Eq.(4).

\begin{equation}
y=f(X\cdot W^{1}+b^{1}) \cdot W^{2}+b^{2}
\label{eq:4}
\end{equation}

In Eq.(4), $X$ represents the input sample, $W^{1}$ and $W^{2}$ are weight matrices, $b^{1}$ and $b^{2}$ are bias vectors, and $f$ is the activation function. The FFNN model is trained using the backpropagation algorithm, which minimizes the loss function by calculating the contribution of each parameter to the error function using the chain rule of differentiation and adjusting the values of each parameter accordingly. For NLP tasks like medical knowledge extraction, FFNN can be used as an upstream network of the CoEx-Bert model to extract features, which are then input into downstream models for further processing. This approach leads to higher quality feature representations, thereby enhancing model performance.

Following the discussion of the FFNN layer, the subsequent section of this paper focuses on the ARC and LN layers. These layers are crucial in deep learning models as they provide mechanisms to enhance the performance and training efficiency of neural networks.

The ARC layer is commonly utilized in deep learning to introduce residual connections in neural networks, thereby facilitating the training of deep models. The main concept involves adding the output of the previous layer to the input of the current layer, forming an additive residual. Mathematically, this is represented by the ARC formula shown in Eq.(5), where the output of the previous layer is denoted as $x$ and the input of the current layer is denoted as $y$. The ARC layer computes the sum of these two quantities, resulting in the output of the current layer. Incorporating the ARC layer within deep neural networks helps mitigate issues such as gradient vanishing and gradient explosion, leading to improved training efficiency and accuracy. For the purpose of this paper, an ARC layer is implemented by introducing a residual block, which consists of two fully connected layers, as depicted in Eq.(6).

\begin{equation}
f(x, y) = x + y
\label{eq:5}
\end{equation}

\begin{equation}
y = f(x, W_1, W_2) = x + W_2\sigma(W_1x + b_1) + b_2
\label{eq:6}
\end{equation}

In Eq.(6), $x$ represents the output of the previous layer, $W_1$ and $W_2$ are the weight matrices of the two fully connected layers, $b_1$ and $b_2$ are their respective bias vectors, and $\sigma$ represents the activation function employed. The residual block takes the output $x$ as its input and generates the output $y$, with the ARC layer represented by the term $x + W_2\sigma(W_1x + b_1)$. The addition of residual blocks and the ARC layer between network layers effectively enhances the representation and training efficiency of deep neural networks.

On the other hand, the LN layer is a normalization technique that operates by normalizing each feature of an individual sample across the feature dimension. It provides a means of standardizing the output of each layer within a neural network, thereby improving the network's ability to accurately represent data. Mathematically, the LN process is described in Eq.(7).

\begin{equation}
y = \gamma\frac{x - \mu}{\sigma} + \beta
\label{eq:7}
\end{equation}

In Eq.(7), $x$ represents the input to the network, $y$ represents the output, $\mu$ and $\sigma$ represent the mean and standard deviation of the input, respectively, while $\gamma$ and $\beta$ are trainable scale and shift factors within the network. LN differs from Batch Normalization in that it performs normalization solely on the current layer, without relying on the batch size. This makes LN more robust in network training. In the present study, LN is employed to normalize the output of each layer in the network, thus improving training stability and accuracy.

\subsection{Shared Parameters in CoEx-BERT}
\noindent

CoEx-BERT incorporates Dropout, Linear, and Sigmoid layers to enhance its functionality. The Dropout layer introduces stochasticity by randomly nullifying components of the input tensor with a defined probability ($p$=0.2). This encourages the model to learn diverse independent features, effectively countering overfitting. The strength of the Dropout technique lies in its ability to reduce co-adaptation amongst neurons, leading to lower generalization error in complex models.

The Linear layer in CoEx-BERT maps feature vectors to their corresponding target vectors through matrix multiplication and addition operations. This process is guided by the model learning optimal weight and bias parameters that align with the dataset structure. The output from the Linear layer is then activated using the Sigmoid function, guaranteeing that the predictions fall within the range of 0 to 1. The Sigmoid function, characterized by an S-shaped curve, demonstrates constant gradients and differentiability, making it suitable for binary classification tasks. Eq.(8) provides the numerical representation of the Sigmoid function.

\begin{equation}
sigmoid(x) = \frac{1}{1+e^{-x}}
\label{eq:8}
\end{equation}

Following the application of the aforementioned model configurations to the extracted features, a joint submodel is constructed. This submodel addresses intimately related tasks that are not entirely identical. While it possesses distinct parameters and activation functions, the submodel shares a common feature extractor, reducing parameter complexity. A crucial joint layer consolidates the features obtained from all subtasks, producing a unified feature vector. This consolidated vector facilitates comprehensive knowledge extraction, enabling a more holistic understanding of the data.

\subsection{Edge Computing Platform Deployment} 
\noindent

To deploy the model on edge devices and optimize it through parameter tuning, a distributed architecture with the necessary software and services is established. For drug knowledge extraction, edge devices are equipped with appropriate NLP libraries and models.

\textbf{Edge Platform Deployment Method}: In edge computing, the first step involves establishing a distributed architecture and deploying the required software and services on edge devices. To facilitate drug knowledge extraction, NLP libraries and models need to be installed on the edge devices.

\textbf{Low latency of edge deployment}:
By deploying the CoEx-BERT model on edge devices, the need to transmit data over large network distances for remote processing is eliminated. This significantly reduces the latency associated with data transport and processing. Situating computation services closer to the data source facilitates rapid access to critical computing resources, enhancing operational efficiency.

\textbf{Considerations for Data Privacy and Security}: As data on edge devices contain sensitive patient information, special attention must be given to data privacy and security in the design of an edge computing architecture. Data should be encrypted during transmission and storage, and access control mechanisms should be implemented to restrict access to sensitive information.

\section{Experiment and Evaluation of Results}
\subsection{Data Processing}
\noindent

This paper employs text recognition technology to extract textual content from the modern medical book 
\verb+"+Zhonghua Bencao, Uyghur Medicine Volume\verb+"+. The extracted texts are manually annotated, with annotations encompassing entity recognition, classification, and sequence labeling tasks. The open-source annotation tool Doccano is used to mark specific content elements in the text, such as herbal names, therapeutic effects, and usage methods. A total of 14,693 triples are annotated in a text dataset of size 1.1M. These marked elements are assigned specific symbols to highlight their significance. The annotation results undergo rigorous review and revision to ensure data accuracy and consistency, resulting in a meticulously crafted dataset that serves as the foundation for subsequent analysis and research.

\subsection{Experimental Setup and Procedure}
\noindent

Standardized parameter configurations are applied, including a weight decay of 0.01, a training batch size of 16, and 128 negative samples generated for each positive sample during training. The Adagrad optimizer is used with a learning rate of $1e^{-5}$ for model parameter optimization. The training process consists of 20 epochs. The model is deployed on Jetson Nano, with a computing power of 0.5 TFLOPs and a GPU configuration of a 128-core NVIDIA Maxwell GPU. The memory parameter is 4GB 64-bit LPDDR4, with a data rate of 24.6GB/s.

The initial model takes three distinct tensors as inputs: $input\_ids$, $input\_mask$, and $segment\_ids$. These three embedding vectors, which are crucial in shaping the CoEx-Bert model's input, are illustrated in Fig.5.  

\begin{figure}[h]
  \centering
  \includegraphics[width=\linewidth]{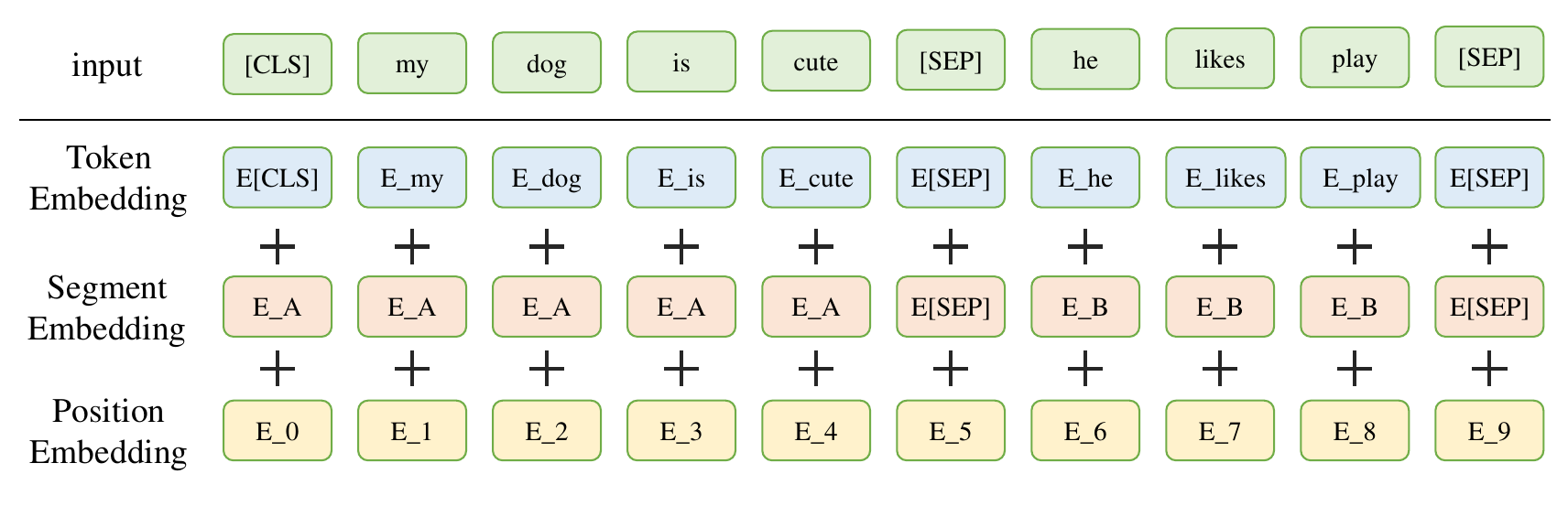}
  \caption{Input and Three Embeddings in CoEx-Bert Model.}
  \label{fig:5}
\end{figure}

\begin{table*}[h]
  \centering
  \parbox[t][0.5cm]{8cm}{\caption{\textbf{Display of Extraction Results.}}}
  \label{tab:2}
  \begin{tabular}{p{5.3cm} p{5.5cm} p{5.5cm}} 
    \toprule
    Example sentence & Prediction results:R & True results:T  \\
    \midrule
    text: The chemical composition of the KunJuTi's seeds includes 45-55\% fat oil, with the main components being lipids (approximately 48\%), linoleic acid (approximately 37\%), palmitic acid, stearic acid, and oleic acid. & R: {(KunJuTi, composition, Stearic acid), (KunJuTi, composition, Fat oil), (KunJuTi, composition, Lipids), (KunJuTi, composition, Palmitic acid), (KunJuTi, composition, Linoleic acid), (KunJuTi, composition, Oleic acid)} & T: {(KunJuTi, composition, Stearic acid), (KunJuTi, composition, Fat oil), (KunJuTi, composition, Lipids), (KunJuTi, composition, Palmitic acid), (KunJuTi, composition, Linoleic acid), (KunJuTi, composition, Oleic acid)}  \\

    text: KaiHeRiBa can be prepared as AiBiKaiHeRiBaXiaoWan, SuFuFeiKaiHeRiBaSan. &  R: {(KaiHeRiBa, preparation, AiBiKaiHeRiBaXiaoWan), (KaiHeRiBa, preparation, AiBiKaiHeRiBaXiaoWan)} & T: {(KaiHeRiBa, preparation, AiBiKaiHeRiBaXiaoWan), (KaiHeRiBa, preparation, AiBiKaiHeRiBaXiaoWan)}  \\
    
    text: OuRuHeSiZiOuZuMi is harvested and used fresh from the fruit cluster or dried in a ventilated drying room into raisins when the fruit matures in autumn. Measures are taken to prevent insect infestation. & R: {(OuRuHeSiZiOuZuMi, storage, Harvest and use the fruit cluster fresh), (OuRuHeSiZiOuZuMi, storage, Dry them into raisins in a drying room for later use.), (OuRuHeSiZiOuZuMi, storage, prevent insect infestation)} &  T: {(OuRuHeSiZiOuZuMi, storage, Harvest and use the fruit cluster fresh), (OuRuHeSiZiOuZuMi, storage, Dry them into raisins in a drying room for later use.), (OuRuHeSiZiOuZuMi, storage, prevent insect infestation)} \\

   \bottomrule
\end{tabular}
 
\end{table*}

In the CoEx-Bert model, the [CLS] symbol plays a significant role as a specialized \verb+"+classification\verb+"+ token within the input sequence. It is responsible for aggregating the embedding vector of the entire input sequence, resulting in a vector suitable for classification tasks. Conceptually, it represents the comprehensive embedding of the entire sentence, which is essential for predictive classification in sentence or text-related tasks. On the other hand, the [SEP] symbol is used to distinguish different segments within the input sequence. When two sentences are concatenated, the CoEx-Bert model inserts the [SEP] symbol to indicate the end of the first sentence and the beginning of the second sentence.

There are three crucial components to the CoEx-Bert input realm. $Token Embeddings$ convert each token into a corresponding vector representation. $Segment Embeddings$ distinguish between two sentences, while $Position Embeddings$ denote the position of each token in the input.

Compared to vectors initialized at random, pre-trained word vectors demonstrate significantly superior performance in entity recognition and relation extraction tasks. These vectors are more proficient at capturing semantic information in words. The extracted features are passed through a Dropout layer, followed by a fully connected layer to yield the output. There is also a fully connected layer after the Dropout layer applied here, where the input size matches the size of the input feature vector or $hidden\_states$. Considering this is a binary classification problem, the output size is 2. Activation is achieved through the Sigmoid function, subsequently squaring the result to obtain the output tensor. The hidden layer parameters are then fed into a second model for parameter sharing.

The second model receives its input as $hidden\_states$ from the output of the first model, setting $batch\_subject\_id$ as the entity's start and end numbers. This model initially computes entity positions in each batch. Based on entity position data, the $hidden\_states$ are split into smaller tensors by batches, extracting the entity segments prior to their broadcast to each document position. Thereafter, the refined $hidden\_states$ are subjected to a secondary transformation via Dropout and a fully connected layer. Given the existence of 16 relations, the classification quantity is set at 16. These stages culminate in the activation through the Sigmoid function, after which the output, as illustrated in Table 2, is generated. Namely, \verb+"+KunJuTi\verb+"+, \verb+"+KaiHeRiBa\verb+"+, and \verb+"+OuRuHeSiZiOuZuMi\verb+"+ are phonetic translations of certain Xinjiang medicines. Similarly, \verb+"+AiBiKaiHeRiBaXiaoWan\verb+"+ and \verb+"+SuFuFeiKaiHeRiBaSan\verb+"+ are the phonetically translated names of medicinal formulations containing components from Xinjiang medicines.

\subsection{Experimental Results}

\subsubsection{Baseline Comparison Result}
\noindent

In the experiment, advanced models were employed as baseline models for the extraction of biomedical information from the Uyghur Medicine dataset. These models encompassed $NovelTagging$, $CopyR$, $GraphRel$, $CopyR_{RL}$, and $CasREL_{LSTM}$.

The findings reveal that the CoEx-Bert model outperformed both $CasREL_{LSTM}$ and other cutting-edge deep learning methodologies in terms of precision, recall, and F1 scores. To be more specific, the CoEx-Bert model achieved remarkable scores of 90.65\% for precision, 92.45\% for recall, and 91.54\% for the F1 score. Consequently, it is evident that the CoEx-Bert model displayed exceptional proficiency in handling the joint extraction task within the context of Uyghur medicinal texts, as illustrated in Table 3.

In summary, the experiments conducted in this paper demonstrate that the CoEx-Bert model outperforms advanced baselines such as $CasREL_{LSTM}$ in the context of joint extraction for Uyghur medicinal texts. This proposed approach exhibits notable improvements in both extraction accuracy and efficiency, which could have substantial implications for this specific task.

Furthermore, the performance of the model varies depending on the amount of training data used. When the same amount of data is used, the CoEx-BERT model outperforms other models. As more labeled data becomes available in the future, it is expected that the performance gap between the CoEx-BERT model and other methodologies will continue to widen. This calls for a more detailed examination of the foundational models and architectural features. 

\begin{table}[h]
  \parbox[t][0.5cm]{8cm}{\caption{\textbf{Evaluation Metric Presentation.}}}
    \label{tab:3}
  \begin{tabular}{lccc} 
    \toprule
    Methods  & Precision & Recall & F1 \\
    \midrule
     $NovelTagging$(Zheng et al.2017)  & 0.624 & 0.317 & 0.420 \\
      $CopyR$(Zeng et al.2018)  & 0.610 & 0.566 & 0.587 \\
      $GraphRel$(Fu et al.2019)& 0.639 & 0.600 & 0.619  \\
      $CopyR_{RL}$(Zheng et al.2019) & 0.779 & 0.672 & 0.721  \\
      $CasREL_{LSTM}$(Wei et al.2020) & 0.842 & 0.830 & 0.836  \\
     \textbf{CoEX-Bert} & \textbf{0.906}  & \textbf{0.924} & \textbf{0.915}  \\
   
   \bottomrule
\end{tabular}

\end{table}

In comparison to sophisticated baseline models like $CasREL_{LSTM}$, the CoEx-BERT model demonstrates superior performance in extracting Uyghur traditional medicine co-extraction. This significantly improves the precision and efficiency of the extraction process. The changes in the CoEx-Bert loss function with training iterations are shown in Fig.6.

\begin{figure}[h]
  \centering
  \includegraphics[width=8.5cm,height = 6cm]{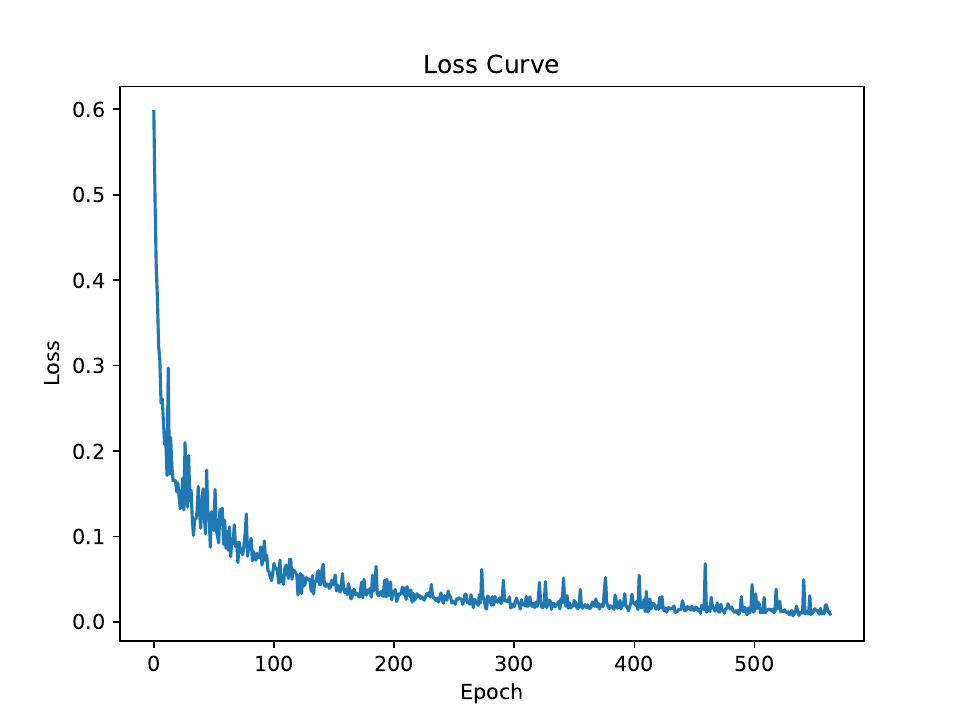}
  \caption{Changes in loss function across training epochs.}
  \label{fig:6}
\end{figure}

This paper confirms, based on the analysis of experimental results, that the superiority of the CoEx-Bert model lies in its enhanced ability to process intra-sentence semantic information. Its pre-training process gives it a stronger language understanding capability. In the context of this experiment, considering the complexity of the Uyghur ethnic medicine attributes, they involve not only the relation between words but also contextual information. The CoEx-Bert model is more suitable for understanding and utilizing this contextual information for the task of extracting medicines.

\subsubsection{Statistical Significance Testing}
\noindent

This paper employed statistical significance testing, specifically the t-test, to determine whether there is a significant difference in the average performance of our developed CoEx-Bert model and the most competitive baseline model $CasREL_{LSTM}$ over multiple experiments. These quantitative analysis methods help us determine if the differences between the two models are large enough to be unlikely to be caused merely by random noise.

In comparison, this paper chose the previously well-performing $CasREL_{LSTM}$ as our baseline model to measure whether substantial progress has been made with our self-developed CoEx-Bert model. We not only conducted single experiments but also carried out five random repeated experiments to provide more reliable and stable evaluation results.

Regarding specific metric comparisons, this paper focused on \verb+"+accuracy\verb+"+, which is a common and intuitive measure representing the proportion of correct predictions made by a model. Table 4 lists our corresponding evaluation results when performing Uyghur drug knowledge extraction tasks.

\begin{table}[h]
\parbox[t][0.5cm]{8cm}{\caption{\textbf{Comparison of Accuracy in Five Experiments.}}}
    \label{tab:4}
  \begin{tabular}{lccccc} 
    \toprule
    Methods  & Trial1 & Trial2 & Trial3 & Trial4 & Trial5\\
    \midrule
      $CasREL_{LSTM}$ & 0.840 & 0.847 & 0.843 & 0.831  & 0.827 \\
     \textbf{CoEX-Bert} & 0.908  & 0.905 & 0.911 & 0.903  &  0.902\\
   
   \bottomrule
\end{tabular}
\end{table}

In this paper, five different experimental iterations were conducted, labeled as Trial 1 to Trial 5, in order to draw more robust conclusions. This method of multiple iterative experiments is a common practice aimed at reducing the impact of random factors and obtaining more stable and reliable results.

Upon initial observation, the CoEx-Bert model appears to outperform the baseline model $CasREL_{LSTM}$ in all experiments. To verify whether this observational conclusion is statistically significant, a t-test was conducted. This statistical significance test was performed using Python's scipy library to determine if the performance difference between the two models exceeds the range of random error.

The result of the t-test gave us a T-statistic value of -16.6888 and a P-value of 1.6808e-07. The P-value represents the probability that our observed results would occur under the null hypothesis (i.e. there is no significant difference in performance between the two models).

Since the obtained P-value of 1.6808e-07 is less than the conventional significance level of 0.05, the null hypothesis was rejected, and the alternative hypothesis was accepted. This means that there is a statistically significant difference in performance between the CoEx-Bert model and $CasREL_{LSTM}$ model, which further emphasizes the superiority of our developed CoEx-Bert model for the Uyghur drug knowledge extraction task.

\subsubsection{Performance Evaluation on Edge Devices}
\noindent

In order to deploy the Uyghur language drug information extraction model described in this article on edge computing devices, the NVIDIA Jetson Nano was selected\upcite{32}. The necessary software and libraries, including the Python interpreter, the PyTorch deep learning framework, the ONNX model conversion tool, and Nvidia's TensorRT inference optimization tool, were installed on the Jetson Nano. To enhance the performance of the model on these devices, the original TorchScript model was converted into a TensorRT model using ONNX as an intermediary. TensorRT is a high-performance library specifically designed for deep learning inference, which significantly improves model execution speed and reduces memory consumption. This makes it highly suitable for the NVIDIA hardware environment.

The code handles necessary processes such as loading models, preprocessing input data when needed, calling models for information extraction, and post-processing the output results. The model was then run on the Jetson Nano, and a series of performance tests were conducted. These tests included measuring inference speed and verifying model accuracy. Based on the test results, some optimizations and fine-tuning were performed to better adapt the model to edge computing environments\upcite{33}. Finally, the model's performance was evaluated using experimental metrics, and the results are presented in Table 5 for reference.

\setlength{\hangindent}{18pt}
\noindent
$\bullet$   \textbf{Latency}: The average response time for model inference.

\setlength{\hangindent}{18pt}
\noindent
$\bullet$   \textbf{Bandwidth Consumption}: The network transmission volume utilized during model inference.

\begin{table}[h]
\parbox[t][0.5cm]{8cm}{\caption{\textbf{Edge Computing Performance.}}}
    \label{tab:5}
  \begin{tabular}{lccc} 
    \toprule
    Methods  & Bandwidth Utilization & Average Latency \\
    \midrule
      $CopyR_{RL}$  & 110M & 70ms  \\
      $CasREL_{LSTM}$  & 120M & 72ms  \\
    CoEx-Bert    & \textbf{100M} & \textbf{65ms}  \\
   
   \bottomrule
\end{tabular}
\end{table}

As shown in Table 5, it can be observed that CoEX-Bert exhibits a bandwidth consumption of 100M and an average latency of 65ms. In comparison to other baseline models such as $CasREL_{LSTM}$ with 120M bandwidth consumption and an average latency of 72ms, or $CopyR_{RL}$ with 110M bandwidth consumption and 70ms average latency, CoEX-Bert shows notable improvements.

Based on the above experimental results, the optimized Uyghur Medicine Extraction model, CoEX-Bert, achieves a latency of 65ms when running on Jetson Nano, which effectively meets the real-time extraction requirements. Edge computing devices achieve low latency and lower bandwidth consumption. This indicates that the model can swiftly process input data without imposing a significant load on network transmission\upcite{34}. This is highly advantageous for real-time Uyghur Medicine Extraction applications, enabling efficient extraction services on edge computing devices.

\subsection{Model Generality Evaluation}
\noindent

\verb+"+Zhonghua Bencao\verb+"+ contains not only a wealth of Uyghur medicinal scrolls but also medicinal scrolls from Tibetan, Mongolian, Dai, Miao, and other ethnic minorities. These medical texts, originating from different ethnic groups, share similarities in terms of language structure, corpus usage, and professional vocabulary. While each text has its own unique features, there are significant similarities among them. Importantly, these differences and similarities do not significantly affect the essence of medicines and data structures. Given the unified characteristics and similar data structures of these sources, the CoEx-Bert technology shows promising prospects for application and potential for promotion.

In this section of the study, the CoEx-Bert model was adopted and fine-tuned moderately. Subsequently, an experimental test for information extraction was conducted on 200 medical texts in Mongolian medicine scrolls. Notably, the accuracy rate of this experiment reached as high as 87.43\%, which not only fully demonstrated the application potential of the CoEx-Bert model on other medicine scrolls with similar characteristics but also verified its excellent generalization performance. In other words, when dealing with similar scroll information, the CoEx-Bert model has outstanding adaptability and precision that meet our set standards. This indirectly proves that our model possesses extensive promotional value in scientific research and practical applications.

\section{Conclusion}
\noindent

The findings presented in this research provide a valuable roadmap for initial inquiries into methods and techniques within the domain of Uyghur Medicine Joint Extraction. The experiment outcomes demonstrate that our proposed method, compared to traditional models, can more simply extract both entities and relations from forecasted label sequences. Our process showcases entity recognition performance comparable to singular subtask performance and superior capabilities in relation to relation and entity extraction. Furthermore, it markedly improves recall rates for overlapping relations and partially resolves issues related to non-contiguity and type confusion. Deploying this knowledge extraction model on edge devices can result in various positive impacts, such as reduced data transmission requirements, provision of real-time responses, and enhanced data security. The Uyghur Medicine Entity and relation Extraction methodology within edge computing can offer accurate, real-time foundational data support for various fields, such as integrated Chinese and Western medicine and the modernization of traditional Chinese medicine.

The CoEx-Bert model underscores a joint extraction method, essentially forming an end-to-end information extraction strategy, which offers significant advantages in managing complex structure extraction tasks like triplets. However, upon observing some misclassified samples, it becomes evident that the accuracy of our model exhibits a certain discrepancy with preexisting named entity recognition models when processing the named entity recognition task of individual entities.

As a result, the plan is to integrate the CoEx-Bert model with some high-performing named entity models in future work, with the goal of enhancing the recognition accuracy of individual entities. This will further boost the overall performance of our model, enabling it to more efficiently exploit its advantages in a range of information extraction scenarios.

\mbox{}
\clearpage
\clearpage
\large

  \end{document}